# An Active Contour Model Driven By the Hybrid Signed Pressure Function

Jing Zhao

College of Mathematics and Statistics,

Chongqing University, Chongqing, 401331, P.R.China

Email: 202106021019@cqu.edu.cn

*Abstract:* Due to the influence of imaging equipment and complex imaging environments, most images in daily life have features of intensity inhomogeneity and noise. Therefore, many scholars have designed many image segmentation algorithms to address these issues. Among them, the active contour model is one of the most effective image segmentation algorithms.This paper proposes an active contour model driven by the hybrid signed pressure function that combines global and local information construction. Firstly, a new global region-based signed pressure function is introduced by combining the average intensity of the inner and outer regions of the curve with the median intensity of the inner region of the evolution curve. Then, the paper uses the energy differences between the inner and outer regions of the curve in the local region to design the signed pressure function of the local term. Combine the two SPF function to obtain a new signed pressure function and get the evolution equation of the new model. Finally, experiments and numerical analysis show that the model has excellent segmentation performance for both intensity inhomogeneous images and noisy images.

# 1. Introduction

Vision is the highest level of human perception and one of the most important ways for humans to obtain information from the outside world. However, as the foundation of human vision, images are not only an objective reflection of external objects, but also an important carrier for human perception and understanding of the world. So, images play the most important role in human perception. Image segmentation is a fundamental task in the field of image processing, which is widely used in fields such as image analysis, computer vision, and medical imaging [18]. The purpose of image segmentation is to divide a given image into several object regions, where each region is similar to a certain feature, namely intensity inhomogeneity, color, and texture [34]. And many scholars have designed a large number of image segmentation algorithms for different applications. Among these numerous algorithms, the Active Contour Model (ACM)[1-17]is one of the most effective image segmentation algorithms. Its advantage is that the model can handle topological changes in the evolution curve. Usually, the evolution curve in the active contour model is represented by a zero level set, and the evolution curve is driven to stop at the object boundary by minimizing the energy functional. According to the different expressions of initial contour curves, active contour models can be divided into two categories: parametric active contour models and geometric active contour models. The parameter active contour model is highly dependent on the position of the initial contour curve, which means that the segmentation results obtained for different initial contour curves will be different, and even incorrect segmentation results may be obtained. In addition, it is also very sensitive to changes in the topological structure of the curve. So, in contrast, the geometric active contour model has become a research hotspot for many scholars, because compared to the parametric active contour model, it can better handle the topological structure changes of curves and solve the problems that the parametric active contour model is difficult to solve. Geometric active contour models can be further divided into two categories: edge-based models [1-8] and region-based models [9-17].

In 1997, Caselles et al.[2]studied the GAC model, which is one of the most classic edge-based active contour models. The idea is to integrate image gradient information and geometric active contours into curve evolution theory and design an edge stopping function. However, it still uses image gradient information to segment objects, which makes it difficult to detect objects from images with noise and weak boundaries. Therefore, the GAC model is usually sensitive to weak edges and noise. Unlike edge-based active contour models, region-based active contour models no longer focus on the gradient information of the image, but are constructed based on formulas that calculate the intensity information of the image inside and outside the evolution curve. Region-based models drive the motion of evolution curves based on the distribution of mean, median, variance, and texture of intensity within the region, thereby better handling noise and weak edges. In 1989, Mumford and Shah proposed the M-S model [11]. The M-S model uses the global intensity difference between the average intensity of the inner and outer regions of the evolution curve to guide the evolution curve towards the object boundary. It can achieve accurate segmentation of images with intensity homogeneity. However, it is not possible to segment images with intensity inhomogeneity. In 2001, Chan and Vese et al. [9]proposed the CV model based on the M-S model. The CV model simplifies the M-S model by assuming that the image is piecewise constant and using the level set method [24]to represent regions and boundaries. The CV model can produce better segmentation results than the M-S model and is easier to implement. However, due to the fact that the CV model is only a global model and the proposed assumption is not feasible in practical life, as the intensity inhomogeneity of images is a real problem, the CV model still cannot effectively segment images with intensity inhomogeneity. In 2008, in order to segment images with intensity inhomogeneity, Li et al. [13]proposed an RSF model based on local information. This model approximates the intensity values inside and outside the active contour in a local area by defining two fitting functions. Thus obtaining the data fitting term, and then obtaining the energy functional based on the data fitting term and the level set regularization term. The RSF model also utilizes the kernel function in the data fitting term to extract local intensity information to drive the motion of the curve, enabling the RSF model to segment images with inhomogeneity intensity . However, this model is a local model, so it heavily relies on the initialization of the level set and is prone to getting stuck in local minima. In 2010, Zhang et al. [38]proposed a new level set model with selective binarization and Gaussian filtering regularization (SBGFRLS). This model utilizes global statistical information from both the external and internal aspects of the evolution curve to design a region based signed pressure (SPF) function, which is then selectively penalized as binary and regularized using a Gaussian filter. The advantage of this method is that it proposes a new region based signed pressure (SPF) function, which can effectively drive the evolution curve to stop at weak or blurred edges of the image. However, due to the fact that the SBGFRLS model is only based on the global information of the image, it can only segment simple intensity inhomogeneious images but cannot effectively segment complex intensity inhomogeneious images. In 2019, Fang et al. [40]proposed a weighted hybrid signed pressure function driven image segmentation active contour model (WHRSPF) based on Zhang's SBGFRLS model. In this model, a global region based SPF function is first designed as the driving center. Secondly, a local region based SPF function is also defined. By combining the SPF function of the global region with the SPF function of the local region, a hybrid signed pressure function is defined, which can improve the segmentation efficiency and accuracy of the model. In 2020, Liu et al. [43]proposed a novel active contour model driven by global and local signed energy pressure functions (GLSEPF). Firstly, by calculating the energy difference between the inside and outside of the evolution curve, a global signed pressure function was designed to improve the robustness of the initial curve. Secondly, by calculating the pixel energy difference within the local neighborhood and introducing a local signed pressure function, this model can segment images with intensity inhomogeneity and noise.

This paper is based on the SBGFRLS model, mainly focus intensity inhomogeneious images and noisy images. In intensity inhomogeneious and noisy images, the intensity values of pixels in their local object (background) areas are mostly not equal, and may even differ greatly. Therefore, if the image segmentation model only relies on global information, it is completely insufficient, Therefore, we designed a signed pressure function that simultaneously extracts global and local information from images, and proposed an active contour model driven by a hybrid signed pressure function. The main research content is as follows:

(1) Starting from the global information of the image, a new global term signed pressure function is introduced by combining the average intensity of the inner and outer regions of the curve and the median intensity of the inner regions of the curve.

(2) At the same time, we further divide the local object area and local background area into two sub regions, extract the average intensity in the sub regions, and use the energy difference between the inner and outer regions of the curve in the local region to design the sign pressure function of the local term.

(3) Combine the two SPF to obtain a new signed pressure function and substitute it into the SBGFRLS model to obtain the evolution equation of the new model.

The rest of the paper is organized as follows: Section 2 introduces the background and the related work of our research, including the CV model and the SBGFRLS model. Section 3 describes our proposed model in detail. Section 4 shows extensive experimental results. Section 5 concludes the paper.

## 2. Related work

### 2.1 CV model

Chan and Vese [9] designed a region-based active contour model, which is an improvement of the M-S model [2]. Set the input image is $I(x):\Omega \to R^+$ and a closed contour curve $C(q):[0,1]\to R^2$ which can divide the image domain into internal region $C_{in}$ and external region $C_{out}$. The energy functional of the CV model can be expressed as:

$$E^{CV}(C,c_1,c_2) = \lambda_1 \int_{C_{in}} |I(x)-c_1|^2 dx + \lambda_2 \int_{C_{out}} |I(x)-c_2|^2 dx + \mu \cdot length(C) \quad (1)$$

Where $\lambda_1$、$\lambda_2$ and $\mu$ are three normal numbers, $c_1$ and $c_2$ are two piecewise constants, representing the average intensity of the internal region $C_{in}$ and external region $C_{out}$, respectively.

In the level set method [24], a closed curve $C$ is a zero level set of a Lipschitz function $\phi(x)$. And it is represented as follows:

$$\begin{cases} \phi(x) > 0 & if(x) \in Inside(C) \\ \phi(x) = 0 & f(x) \in On(C) \\ \phi(x) < 0 & f(x) \in Outside(C) \end{cases} \quad (2)$$

To minimize the energy functional, replace the curve $C$ with a level set function, and the energy functional of the CV model can be rewritten as

$$E^{cv}(\phi(x),c_1,c_2) = \lambda_1 \int_{\Omega} |I(x)-c_1|^2 H(\phi(x))dx + \lambda_2 \int_{\Omega} |I(x)-c_2|^2 (1-H(\phi(x)))dx$$

$$+ \mu \cdot \int_{\Omega} \delta(\phi(x))|H(\phi(x))|dx \quad (3)$$

Where $I(x)$ is the input image, $\lambda_1, \lambda_2 > 0$ and $\mu > 0$ is the weighted parameter, $\phi(x)$ represents the level set function, $H(\phi(x))$ and $\delta(\phi(x))$ are the Heaviside function and the Dirac function, respectively. Their functions are defined as follows:

$$H(\phi) = \frac{1}{2}\left[1 + \frac{2}{\pi}\arctan(\frac{\phi}{\varepsilon})\right] \quad (4)$$

$$\delta(\phi(x)) = [H(\phi(x))]' = \left[\frac{1}{2}(1+\frac{2}{\pi}\arctan(\frac{\phi}{\varepsilon}))\right]' = \frac{1}{\pi}\frac{\varepsilon}{\varepsilon^2+\phi^2} \quad (5)$$

Using the alternating iteration algorithm to minimize the energy functional (3), first fix the level set function $\phi(x)$, and obtain the minimum value point for variables $c_1$ and $c_2$ as follows:

$$c_1 = \frac{\int_\Omega I(x) \cdot H(\phi(x))dx}{\int_\Omega H(\phi(x))dx} \quad (6)$$

$$c_2 = \frac{\int_\Omega I(x) \cdot (1-H(\phi(x)))dx}{\int_\Omega (1-H(\phi(x)))dx} \quad (7)$$

Clearly, $c_1$ and $c_2$ represent the average intensity values of pixels in the inner and outer regions of the active contour curve, respectively. Fix $c_1$ and $c_2$, then use variational and gradient descent methods to obtain the minimum point of the energy functional (3) with respect to the level set function $\phi(x)$, which is the steady-state solution of the evolution equation:

$$\frac{\partial \phi}{\partial t} = -\delta(\phi(x))(\lambda_1(I(x)-c_1)^2 - \lambda_2(I(x)-c_2)^2) + \mu\delta(\phi(x))div(\frac{\nabla\phi}{|\nabla\phi|}) \quad (8)$$

The CV model is based on global information for modeling, thus enabling image segmentation. However, since the CV model assumes that the image intensity is piecewise constant distributed from the beginning, it is not effective for handling images with inhomogeneity intensity .

**2.2 SBGFRLS model**

Subsequently, in order to combine the advantages of the GAC model and the CV model, Zhang et al. proposed an active contour model driven by a signed pressure function [38], abbreviated as the SBGFRLS model. Based on the intensity difference of the input image and the average intensity of the inner and outer regions of the evolution curve, constructing an SPF function can effectively segment intensity inhomogeneious images. The definition of the SPF function is:

$$spf(I(x)) = \frac{I(x)-\frac{c_1+c_2}{2}}{\max\left(\left|I(x)-\frac{c_1+c_2}{2}\right|\right)}, \quad x \in \Omega \quad (9)$$

Where the two constants $c_1$ and $c_2$ are respectively represented as the average intensity of the inner and outer regions of the curve. The calculation formula is the formula (6) and (7).The author replaced the edge stopping function (ESF) in the GAC model with the SPF function. The evolution equation is obtained as follows:

$$\frac{\partial \phi}{\partial t} = spf(I(x)) \cdot \left( div\left(\frac{\nabla \phi}{|\nabla \phi|}\right) + \alpha \right) |\nabla \phi| + \nabla spf(I(x)) \cdot \nabla \phi, \quad x \in \Omega \quad (10)$$

Where $\alpha$ is the speed parameter that controls the update of the level set function, and $\nabla \phi$ is the gradient of the level set function.

In the SBGFRLS model, the level set function can be initialized to constants, which have different signs inside and outside the contour.This is very simple to implement in practice. In the traditional level set methods, the curvature-based term $div(\nabla \phi/|\nabla \phi|)|\nabla \phi|$ is usually used to regularize the level set function $\phi$. Since $\phi$ is an SDF that satisfies $|\nabla \phi| = 1$, the regularized term can be rewritten as $\Delta \phi$, which is the Laplacian of the level set function $\phi$. Because the evolution of a function with its Laplacian is equivalent to a Gaussian kernel filtering the initial condition of the function. Thus we can use a Gaussian filtering process to further regularize the level set function. The standard deviation of the Gaussian filter can control the regularization strength. Since the model utilize a Gaussian filter to smooth the level set function to keep the interface regular, the regular term $div(\nabla \phi/|\nabla \phi|)|\nabla \phi|$ is unnecessary. In addition, the term $\nabla spf(I(x)) \cdot \nabla \phi$ in Eq. (10) can also be removed, because our model utilizes the statistical information of regions, which has a larger capture range and capacity of anti-edge leakage. Finally, the level set formulation of the proposed model can be written as follows:

$$\frac{\partial \phi}{\partial t} = spf(I(x)) \cdot \alpha |\nabla \phi|, \quad x \in \Omega \quad (11)$$

The signed pressure function proposed in the SBGFRLS model is capable of adjusting curve motion based on the intensity values of the inner and outer regions of the active contour. The signed pressure function guides the evolution curve to shrink when the active contour is outside the segmentation object, and it in turn guides the evolution curve to expand when the active contour is inside the segmentation object, This design can make the contour curve movement flexible and the segmented image effect good. However, since the model is based on the global information of the image, the SBGFRLS model can only handle images with relatively simple intensity information, and the effect on processing images with complex intensity information is not ideal.

# 3. The proposed model

**3.1 The design of global SPF function**

The SBGFRLS model [38]is available for segmenting images with intensity inhomogeneity by calculating the global information of the images as the driving center. However, in the real world, the foreground and background regions of images are usually inhomogeneous, and if only the global information of the image is utilized, the segmentation is often not accurate enough. Therefore, in order to address this issue, we introduce a global region-based SPF function and a local region-based SPF function, and propose an active contour model driven by a hybrid signed pressure function. The experiment shows that the model can effectively segment images with intensity inhomogeneity and noise.

In the SBGFRLS model, only the global average intensity information of the image is extracted. In order to segment the image more accurately, we consider the global average intensity information and median intensity information of the image. By combining the average intensity of the inner and outer regions of the curve with the median intensity of the inner region of the curve, we obtain a new signed pressure function. The pressure function of this signed is defined as:

$$spf_G(I(x)) = \frac{I(x) - \frac{(c_1 - 2m)^2 - c_2^2}{2(c_1 - c_2)}}{\max\left(\left|I(x) - \frac{(c_1 - 2m)^2 - c_2^2}{2(c_1 - c_2)}\right|\right)}, \quad x \in \Omega \tag{12}$$

Where $c_1$ and $c_2$ are the average intensity of the inner and outer regions of the curve defined in equations (6) and (7), $m$ representing the median intensity within the evolution curve region. These three variables are defined as follows:

$$\begin{cases} c_1 = mean\{I(x) \mid \phi(x) \geq 0, x \in \Omega\} \\ c_2 = mean\{I(x) \mid \phi(x) < 0, x \in \Omega\} \\ m = median\{I(x) \mid \phi(x) \geq 0, x \in \Omega\} \end{cases} \tag{13}$$

Where $I(x)$ is the pixel value of the input image at $x$.

The following is an explanation of the $spf_G$ function in equation (12). Assuming that the intensity of the object area and background area are represented as $c_o$ and $c_b$, respectively, $c_o > c_b$. $C$ represents the evolution curve in a given image. The four stopping positions of the evolution curve are shown in Figure 1:

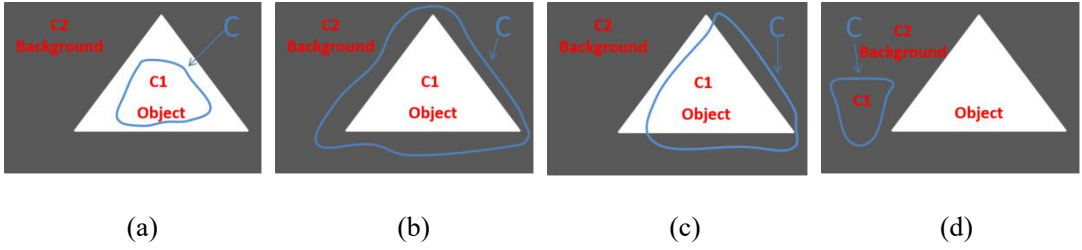

(a)          (b)          (c)          (d)

Figure 1 Four stopping positions of the evolutionary curve

In Figure 1(a), the evolution curve $C$ is completely located within the object region. At this time, the average intensity of the external area of the curve is greater than the average intensity of the background area, and the median intensity of the internal area of the curve is equal to the average intensity of the internal area of the curve, which is also equal to the average intensity of the object area( $m = c_o = c_1, c_2 > c_b$ ). So, the function in equation (12) will be rewritten:

$$I^G = \frac{(c_1 - 2m)^2 - c_2^2}{2(c_1 - c_2)} = \frac{c_1 + c_2}{2} \tag{14}$$

It can be seen that $c_2 < (c_o + c_b)/2 < (c_1 + c_2)/2 < c_1$, this will drive the evolution curve to move towards the background region.

On the contrary, in Figure 1(b), the object region is included within the evolution curve. So the average intensity of the object area is greater than the average intensity of the internal area of the evolution curve, that is $c_o > c_1$, the median intensity of the internal area of the evolution curve is not lower than the average intensity of the external area of the evolution curve, that is $m \geq c_2$. Therefore, there are $c_2 < \left((c_1 - 2m)^2 - c_2^2\right)/2(c_1 - c_2) < (c_o + c_b)/2 < c_1$. So this will drive the evolution curve to shrink towards the object area. In Figure 1(c), the situation is similar to Figure 1(b), so we will not discuss it anymore.

In Figure 1(d), the evolution curve is completely located in the background region. At this time, the average intensity of the object area is greater than the average intensity of the internal area of the evolution curve, and the median intensity of the internal area of the evolution curve is equal to the average intensity of the background area, that is $c_o > c_1$ and $m = c_b$, then there is:

$$c_2 < \left((c_1 - 2m)^2 - c_2^2\right)/2(c_1 - c_2) < (c_o + c_b)/2 < c_1$$

This can drive the evolution curve to move towards the object boundary.

## 3.2 The design of local SPF function

In the local area, which is the black circle as shown in Figure 2, $x$ is a point on the evolution curve, which is the yellow pentagram in Figure 2. The point $y$ is an independent spatial variable, $\Omega_x$ representing a local area with a center $x$ and a radius $r$. That is $\Omega_x = \{|y-x| \leq r, y \in \Omega\}$, the evolution curve $C$ divides the black circle in Figure 2 into two regions, corresponding to the inner and outer regions of the evolution curve. That is, the local area $\Omega_x$ includes the internal area of the curve $\Omega_1$ and the external area of the curve $\Omega_2$, respectively represented as follows:

$$\Omega_1 = \{|y-x| \leq r, y \in (\Omega_x \cap \phi(y) > 0)\}$$
$$\Omega_2 = \{|y-x| \leq r, y \in (\Omega_x \cap \phi(y) < 0)\}$$

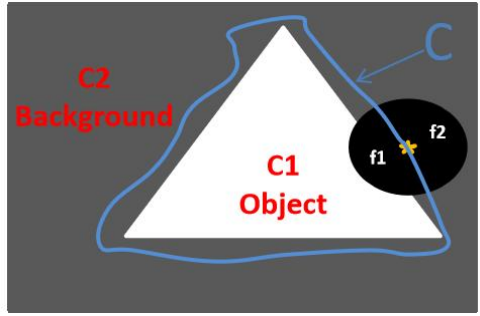

Figure 2 The motion of an evolutionary curve in a local region

That is to say, $f_1$ represents the average intensity value of the internal area divided by the evolution curve of the black circle in the figure, and $f_2$ represents the average intensity value of the external area divided by the evolution curve of the black circle in the figure. The formulas for both are expressed as follows:

$$\begin{cases} f_1 = mean\{I(y) \mid y \in \Omega_1, \Omega_1 = \{\Omega_x \cap (\phi(y) > 0)\}\} \\ f_2 = mean\{I(y) \mid y \in \Omega_2, \Omega_2 = \{\Omega_x \cap (\phi(y) < 0)\}\} \end{cases} \quad (15)$$

And the calculation formula is:

$$f_i(x) = \frac{K_\sigma(x) * \left[M_i^\varepsilon(\phi(x))I(x)\right]}{K_\sigma(x) * M_i^\varepsilon(\phi(x))}, \quad i=1,2 \quad (16)$$

Where $M_1(\phi) = H(\phi)$, $M_2(\phi) = 1 - H(\phi)$ and Gaussian kernel function is represented as:

$$K_\sigma(x) = \frac{1}{(2\pi)^{n/2}\sigma^n} e^{-|x|^2/2\sigma^2} \quad (17)$$

For points $x$, in local regions, the energy inside evolution curve $e_1^l(x)$ and outside evolution curve $e_2^l(x)$ can be represented as follows:

$$e_1^l(x) = \int_{\Omega_1} (I(y) - f_1(x))^2 dy \qquad e_2^l(x) = \int_{\Omega_2} (I(y) - f_2(x))^2 dy \qquad (18)$$

the kernel function $k$ with the above localization property that $k(x-y)$ takes larger values at the points $y$ near the center point $x$, and it decreases to 0 as $y$ goes away from $x$. Therefore, in order to extract local information from the image, the above two local energy $e_1^l(x)$ and $e_2^l(x)$ can be replaced with the following formulas:

$$\begin{cases} e_1^l(x) = \int_{\Omega_x} k_\sigma(x-y)(I(y) - f_1(x))^2 dy \\ e_2^l(x) = \int_{\Omega_x} k_\sigma(x-y)(I(y) - f_2(x))^2 dy \end{cases} \qquad (19)$$

Where $k_\sigma(x-y)$ is the Gaussian kernel function with variance $\sigma$. In the level set function [24], the contour curve $C \in \Omega$ can be represented by the zero level set of a Lipschitz function $\phi : \Omega \to R$, and equation (19) can be written as:

$$\begin{cases} e_1^l(x) = \int_{\Omega_x} k_\sigma(x-y)(I(y) - f_1(x))^2 H(\phi(y)) dy \\ e_2^l(x) = \int_{\Omega_x} k_\sigma(x-y)(I(y) - f_2(x))^2 (1 - H(\phi(y))) dy \end{cases} \qquad (20)$$

Where $H(\phi(y))$ is the Heaviside function. Correspondingly, in the entire image domain, two local energy functions $E_1^l(I(x))$ and $E_2^l(I(x))$ can be written as follows:

$$\begin{cases} E_1^l(I(x)) = \int_\Omega e_1^l(x) dx = \int_\Omega \int_{\Omega_x} k_\sigma(x-y)(I(y) - f_1(x))^2 H(\phi(y)) dy dx \\ E_2^l(I(x)) = \int_\Omega e_2^l(x) dx = \int_\Omega \int_{\Omega_x} k_\sigma(x-y)(I(y) - f_2(x))^2 (1 - H(\phi(y))) dy dx \end{cases} \qquad (21)$$

Therefore, in a local region, by evolving the internal and external information of the curve, a signed pressure function containing local information can be defined as follows:

$$spf_L(I(x)) = \frac{E_2^l(I(x)) - E_1^l(I(x))}{\max(|E_2^l(I(x)) - E_1^l(I(x))|)}, \quad x \in \Omega \qquad (22)$$

Next, explain the meaning of the function $spf_L$ in equation (22). Assuming that in the local area, the average intensity of the object area and background area are $c_o$ and $c_b$, respectively, $c_o > c_b$. The direction of movement of a point $x$ on the evolution curve depends on the difference between the two local energies $e_1^l(x)$ and $e_2^l(x)$. The difference between the two local energies $e_1^l(x)$ and $e_2^l(x)$ in equation (19) can be rewritten as follows:

$$\Delta e^l(x) = e_2^l(x) - e_1^l(x) = \int_{\Omega_x} k_\sigma(x-y)[(I(y) - f_2(x))^2 - (I(y) - f_1(x))^2] dy$$

$$= 2\int_{\Omega_x} k_\sigma(x-y)\left(I(y) - \frac{f_1(x)+f_2(x)}{2}\right)(f_1(x)-f_2(x))dy \quad (23)$$

In equation (23), since the average intensity values inside and outside the evolution curve satisfy $f_1 > f_2$, the direction of the driving force of the point $x$ depends on

$$I(y) - \frac{f_1(x)+f_2(x)}{2}, \quad x \in \Omega, \quad y \in \Omega_x \quad (24)$$

This is the intensity difference between the intensity of the input image at the point and the average intensity of the internal and external regions of the evolution curve.

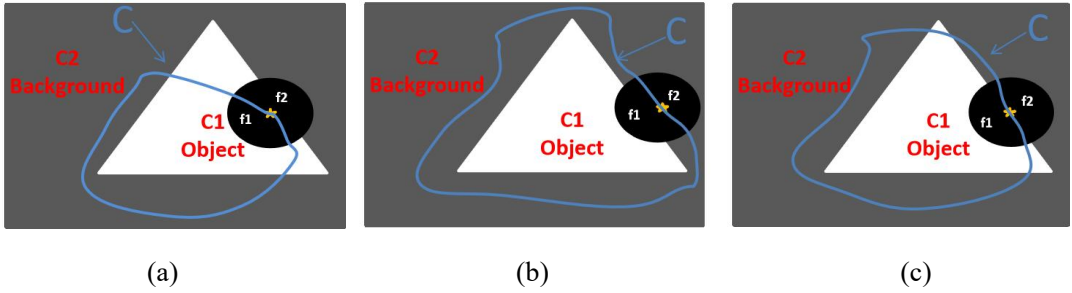

(a)          (b)          (c)

Figure 3 Three stopping positions of the evolutionary curve in a local region

For Figure 3(a), the point $x$ is located within the object area. If the average intensity of the inner region of the curve is close to the average intensity of the object region ($f_1 \approx c_o$), and the average intensity of the outer region of the curve is greater than the average intensity of the background region ($f_2 > c_b$), then there is $c_b < (c_o+c_b)/2 < (f_1+f_2)/2 < c_o$, which drives the evolution curve to move towards the background region.

On the contrary, for Figure 3(b), the point $x$ are located outside the object area. The average intensity of the external area of the curve is close to the average intensity of the background area ($f_2 \approx c_b$), while the average intensity of the internal area of the curve is smaller than the average intensity of the object area ($f_1 < c_o$), then there is $c_b < (f_1+f_2)/2 < (c_o+c_b)/2 < c_o$. Therefore, the evolution curve will move towards the object area.

For Figure 3(c), the point $x$ is located on the boundary of the object. When the contour curve reaches the boundary, the average intensity of the external area of the evolution curve is $(c_o+c_b)/2 \approx (f_1+f_2)/2$, and the contour curve will stop at the object boundary. Similarly, when the average intensity of the local object area is smaller than the average intensity of the local background area ($c_o < c_b$), the direction of point $x$ motion is consistent with the above discussion.

### 3.3 Implementation

Two signed pressure functions based on global information and local information were proposed in section 3.1 and 3.2. Global information helps the evolution curve to be closer to the object boundary, while local information is helpful in describing the changes in local intensity of the object image. Then, the two signed pressure functions are linearly convex combined to construct a signed pressure function that integrates global and local information. The new signed pressure function is obtained as follows:

$$spf_Z(I(x)) = w(spf_G(I(x))) + (1-w)spf_L(I(x)), \quad x \in \Omega \quad (25)$$

Where $w$ is a fixed constant that adjusts the local and global SPF functions. And the values of $w$ and $1-w$ are both between 0 and 1, used to adjust the proportion between local and global terms. That is, when the ratio $w$ is larger, the global term plays the main role; when the ratio $w$ is smaller, indicating that the local term plays the main role.

We substitute the new signed pressure function equation (25) into the level set evolution equation (13) of the SBGFRLS model, and obtain the evolution equation of the new model as follows, and define the name of the new model in this paper as the HZSPF model.

$$\frac{\partial \phi}{\partial t} = spf_Z(I(x)) \cdot \alpha |\nabla \phi|$$

$$= \alpha [w(spf_G(I(x))) + (1-w)spf_L(I(x))] \cdot |\nabla \phi|, \quad x \in \Omega \quad (26)$$

The evolution equation (26) is numerically solved using the finite difference method. The main idea of the finite difference method is to use the ratio of the function difference between adjacent points to the distance between those two points to represent the derivative of the function with respect to their respective variables. Firstly, define the coordinate representation of pixel $x$ in the two-dimensional plane as $(a,b)$, with a time step $\Delta t$. Let $\phi_{i,j}^n = \phi(a_i, b_j, n\Delta t)(n \geq 0 且 1 \leq i \leq M, 1 \leq j \leq N)$ be an approximate value of $\phi(a,b,t)$, where $M \times N$ represents the size of the image. If the forward difference is used for the time partial derivative $\partial \phi / \partial t$, and the center difference is used for the spatial partial derivative $\partial \phi / \partial a$ and $\partial \phi / \partial b$, the initial condition $\phi_{i,j}^0 = \phi_0$ for solving the level set function is set to $n=0$, then there is:

$$\Delta_a \phi_{i,j}^n = \frac{\phi_{i+1,j}^n - \phi_{i-1,j}^n}{2h}, \quad \Delta_b \phi_{i,j}^n = \frac{\phi_{i,j+1}^n - \phi_{i,j-1}^n}{2h}, \quad n = 0,1,2,... \quad (27)$$

The spatial step size can be set $h = \Delta a = \Delta b$, assume that $h=1$ and the above equation

becomes:

$$\Delta_a \phi_{i,j}^n = \frac{\phi_{i+1,j}^n - \phi_{i-1,j}^n}{2}, \quad \Delta_b \phi_{i,j}^n = \frac{\phi_{i,j+1}^n - \phi_{i,j-1}^n}{2}, \quad n = 0,1,2,... \quad (28)$$

So the evolution equation (26) becomes

$$\frac{\phi_{i,j}^{n+1} - \phi_{i,j}^n}{\Delta t} = spf_Z(I(x)) \cdot \alpha \cdot \sqrt{\left(\Delta_a \phi_{i,j}^n\right)^2 + \left(\Delta_b \phi_{i,j}^n\right)^2 + \eta}, \quad n = 0,1,2,... \quad (29)$$

$spf_Z(I(x))$ As shown in equation (25), $\eta$ is any constant greater than zero and very small, used to avoid singularity problems in the above equation. The specific algorithm summary of the HZSPF model is as follows:

---

**Algorithm 1**

---

| | |
|---|---|
| Step1: | Input the initial image and set parameters: $\eta$、$\mu$、$\nu$, the size of the local window $k$ and standard deviation $\sigma$、$\alpha$、$\omega$ and the maximum number of iterations $N$ |
| Step2: | Initialization $\phi_0$ and $n=1$ |
| Step3: | Update $c_1(\phi)$、$c_2(\phi)$ and $m$ by Eq.(6)、Eq.(7) and Eq.(13) |
| Step4: | Update $f_1(\phi)$ and $f_2(\phi)$ by Eq.(16) and update $E_1^l(I(x))$ and $E_2^l(I(x))$ by Eq.(21) |
| Step5: | Update $\phi(x)$ by Eq.(26) |
| Step6: | Test whether the evolution of the level set function curve satisfies the convergence condition or reaches the maximum number of iterations $N$, otherwise return to Step 2 |

---

The convergence condition of the model in this chapter is $\left\|\frac{\phi_{i,j}^{n+1} - \phi_{i,j}^n}{\phi_{i,j}^n}\right\|_\infty < T$, where $T = 0.01$。

## 4. The experimental results and analysis

In this section, we use the proposed model to perform segmentation experiments on single-object intensity inhomogeneious images, medical images, multi-object intensity inhomogeneious images, and different types of noisy images. The computer configuration for all experiments in this paper is: Windows 10 operating system, Inter (R) Core (TM) i5-13500HX CPU, 16GB RAM. The models were tested in MATLAB 2018a environment, and all images were not processed with prior information to ensure consistency in the experiment.

In order to objectively and quantitatively evaluate the segmentation performance of various models on images, this paper adopts two evaluation indicators, namely Dice Similarity Coefficient [41] (DSC) and Jaccard Similarity Coefficient [42] (JS). The defined formulas are as follows:

$$DSC = \frac{2N(S_1 \cap S_2)}{N(S_1) + N(S_2)} \tag{30}$$

$$JS = \frac{N(S_1 \cap S_2)}{N(S_1 \cup S_2)} \tag{31}$$

In the above two equations, $S_1$ and $S_2$ respectively refer to the object area of the original image standard and the experimental object area obtained after model segmentation, $N(\cdot)$ represents the set of pixels contained in the corresponding area. The range of DSC and JSC values is [0,1]. If the values of DSC and JS are closer to 1, it indicates that the model has a better segmentation effect on the image and achieves more accurate segmentation.

**4.1 Segmentation of images with intensity inhomogeneity**

**(1)Verify the segmentation performance of the HZSPF model on images with single-object intensity inhomogeneity.** Three test images were selected, as shown in Figure 4. The first row of test images has obvious edge blurring and inhomogeneious intensity, and the complexity of image details varies among the three images. The second row of images displays the initial contours of the test images, while the third row shows the segmentation results obtained using the model in this chapter. From the experimental results, it can be analyzed that the proposed HZSPF model performs well in segmenting images with intensity inhomogeneity, and is suitable for this type of image segmentation. It can eliminate the interference of weak boundaries and complex intensity information changes, accurately detect the contour of the object, and successfully extract the object.

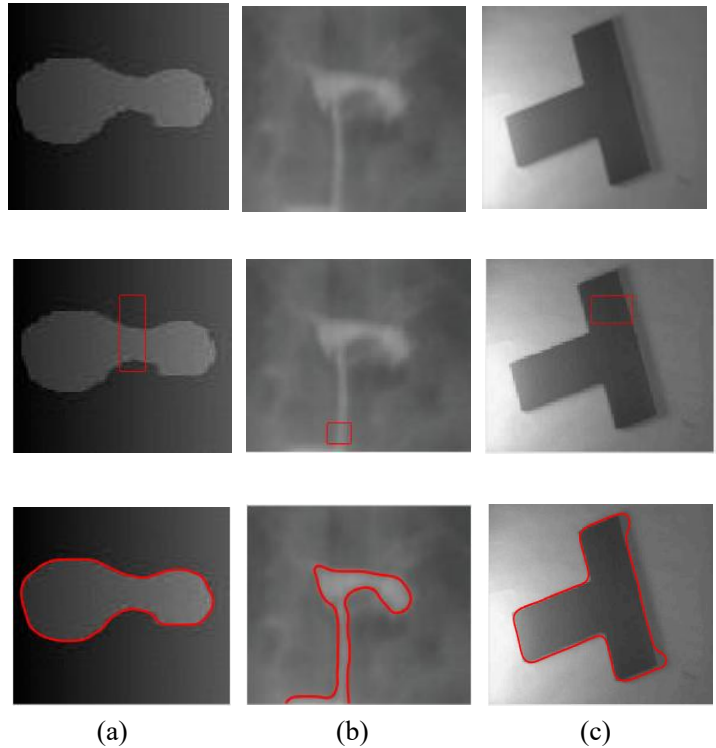

(a) (b) (c)

Figure 4 The segmentation results of the HZSPF model on the images with single-object intensity inhomogeneity

**(2)Verify the segmentation performance of the HZSPF model on medical images with intensity inhomogeneity.** Firstly, four different types of medical images were selected for testing, including ultrasound, magnetic resonance imaging, and contrast imaging. Medical images are different from other types of images. Due to the interference of magnetic fields between devices and many uncontrollable external factors, most medical images are blurry, with unclear object boundaries and easily affected by noise interference. Therefore, compared to the image with single-object intensity inhomogeneity in Experiment 1, the segmentation difficulty of this type of image is greater. As shown in Figure 5, the first row is the test image, and it can be seen that the complexity of various medical images is inconsistent. The second row of images displays the initial contours set for the four test images, while the third row shows the segmentation results obtained using the model in this chapter. It can be clearly observed from the figure that the HZSPF model in this chapter is suitable for medical image segmentation and has good segmentation results.

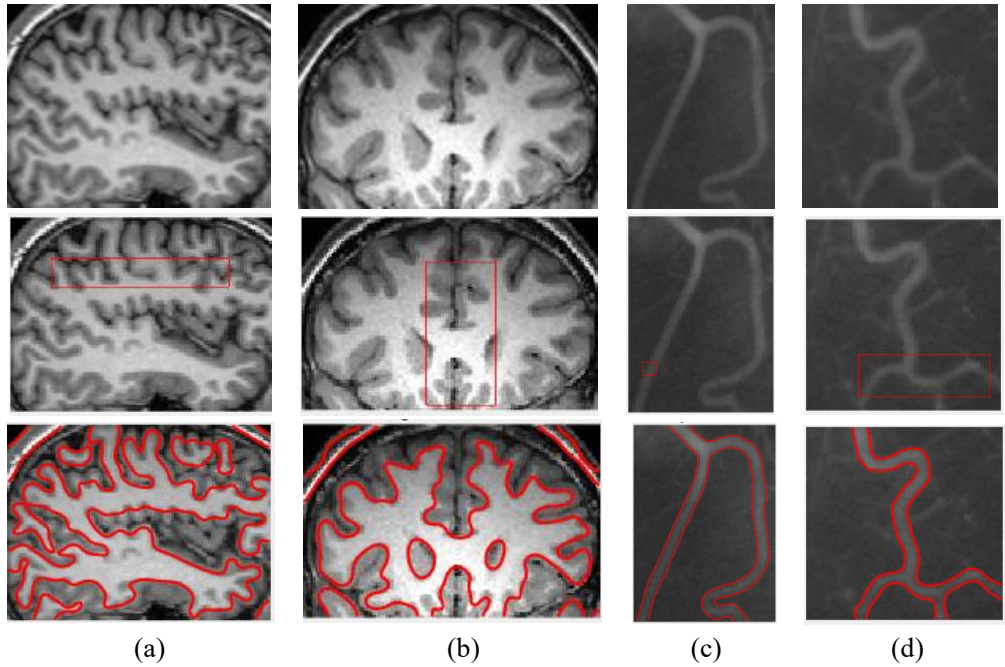

(a) (b) (c) (d)

Figure 5 The segmentation results of the HZSPF model on the medical images

**(3)Verify the segmentation performance of the HZSPF model on images with multi-object intensity inhomogeneity.** As shown in Figure 6, three multi-object images with intensity inhomogeneity were selected, among which the first image has two objects, while the second and third images both have three objects.

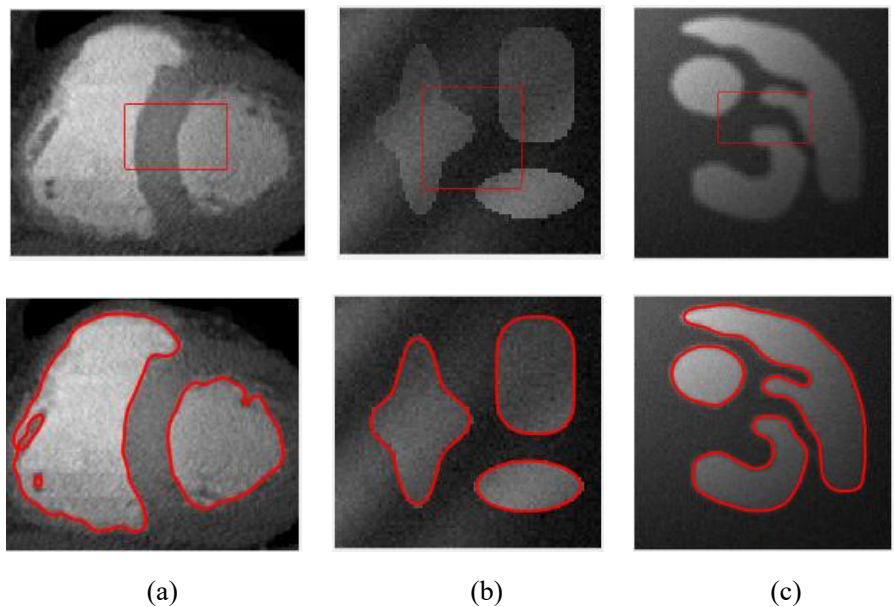

(a) (b) (c)

Figure 6 The segmentation results of the HZSPF model on the images with multi-object intensity inhomogeneity

In the figure 6, the first row shows the initial contours of the three images, while the second row shows the segmentation results of the model for the three images in this chapter. It can be seen that whether it is 2 or 3 objects, the HZSPF model can successfully segment even in the case of intensity inhomogeneity, which proves the feasibility of the model in segmenting multi-object intensity inhomogeneity images.

**4.2 Segmentation of images with noise**

**(1)Conduct experiments on noise images of the same type but with varying level.** As shown in Figure 7, the first to fourth rows represent two images contaminated by Gaussian noise with varying degrees of mean zero and noise variances of 0.01, 0.02, 0.03, and 0.04, respectively. And the first and third columns represent the initial contours of two images with varying degrees of Gaussian noise, while the second and fourth columns represent the segmentation results of the HZSPF model on the two noisy images, respectively. It can be seen that the HZSPF model has a very good segmentation effect on noisy images with different degrees.

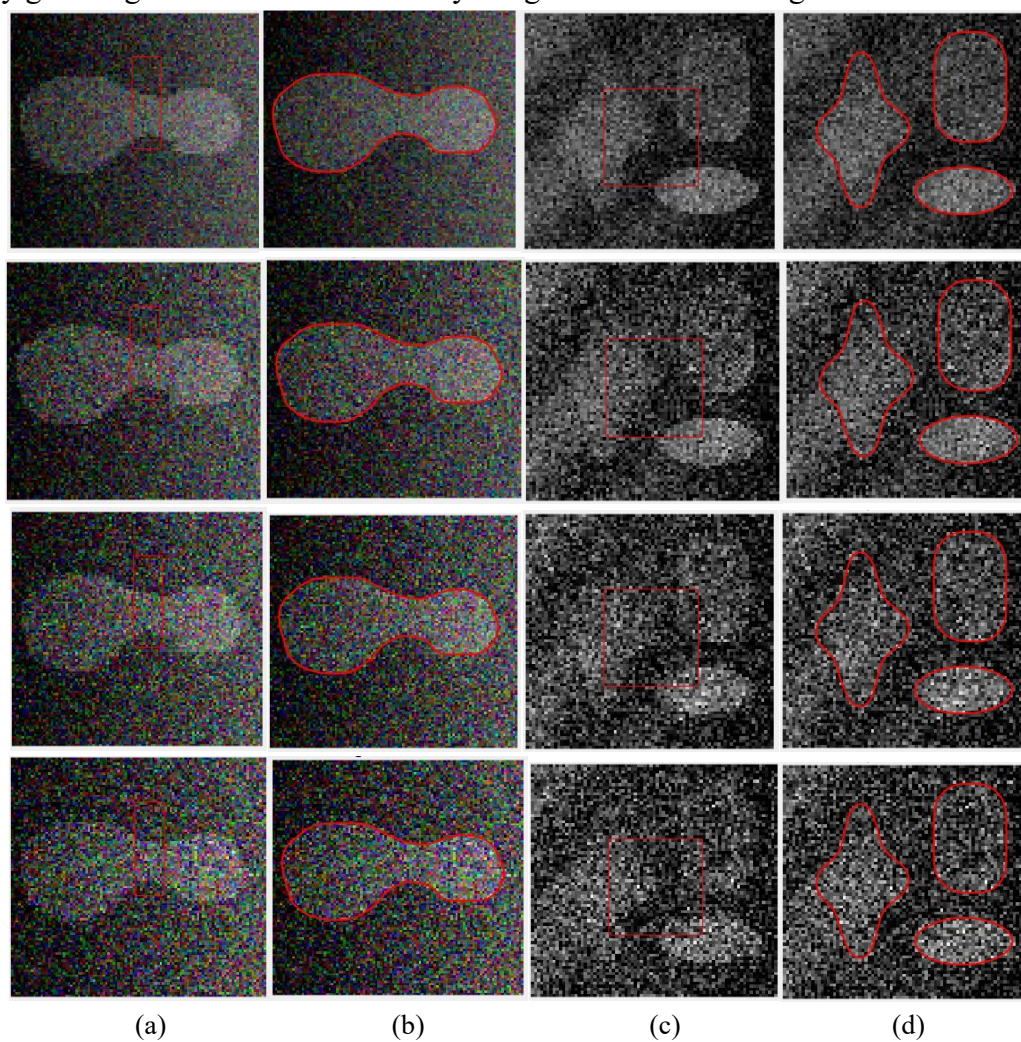

(a)　　　　　　　(b)　　　　　　　(c)　　　　　　　(d)

Figure 7 The segmentation results of the HZSPF model under different levels of Gaussian

noises

Using the groundtruth image and equations (30) and (31), calculate the DSC and JS values of the HZSPF model for segmentation of different levels of Gaussian noises, as shown in Table 1.

Table 1 DSC and JS values of segmentation results in Figure 7

|     | $\sigma_1(0.01)$ | $\sigma_2(0.02)$ | $\sigma_3(0.03)$ | $\sigma_4(0.04)$ |
|-----|------------------|------------------|------------------|------------------|
| DSC | 0.9857           | 0.9781           | 0.9738           | 0.9732           |
| JS  | 0.9884           | 0.9776           | 0.9743           | 0.9662           |

Draw the data in Table 1 into a line chart, and observe from the chart that the trend of the line changes gradually decreases. As shown in Figure 8, it indicates that as the noise variance increases, the boundaries of the object will be affected and become more blurred, which does indeed affect segmentation performance. However, from the data in the table, it can be seen that both DSC and JS values are very close to 1, and the degree of weakening in segmentation performance changes very little. This proves that the HZSPF model is still relatively stable when dealing with different levels of Gaussian noise and can accurately segment objects.

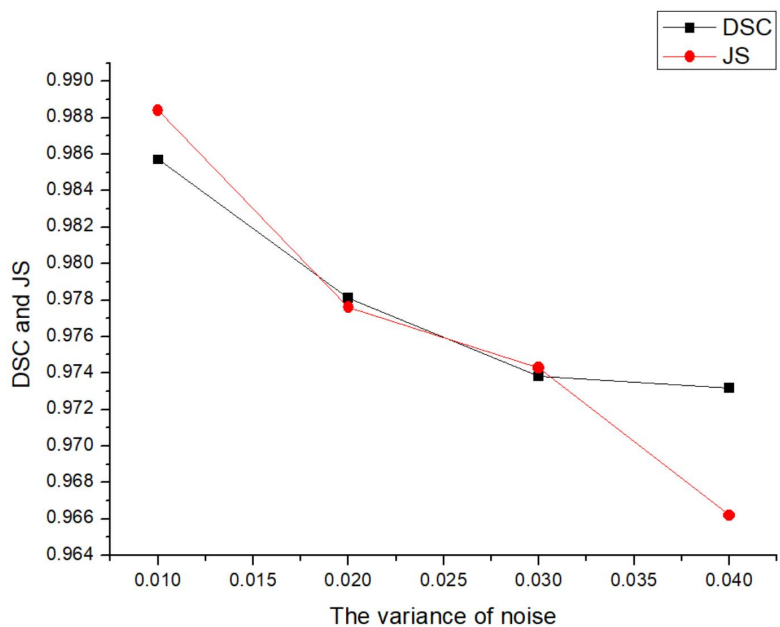

Figure 8 The broken line graph of DSC and JS values in Table 1

**(2)Experiment on different types of noisy images.** In Figure 9, the first column shows the segmentation results of Gaussian noise images with a mean of 0 and a variance of 0.01. The second column shows the segmentation results of salt and pepper noise images with a noise density of 0.01. The third column shows the segmentation results of Poisson noise images. The fourth column shows the segmentation results of speckle noise images with a mean of 0 and a variance of 0.01. From the segmentation results, it can be seen that the model proposed in this paper can accurately segment images even under different types of noise pollution.

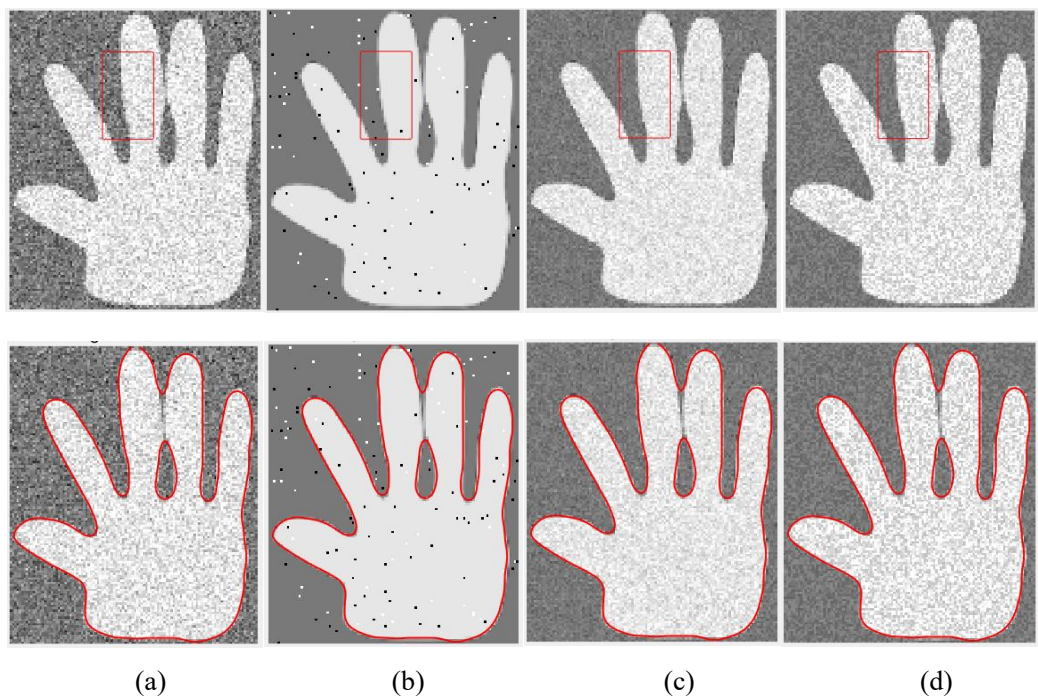

(a)          (b)          (c)          (d)

Figure 9 The segmentation results of the HZSPF model under different noises

Using the groundtruth image and equations (30) and (31), calculate the DSC and JS values of the HZSPF model for segmentation of four different types of noisy images, as shown in Table 2.

Table 2 DSC and JS values of segmentation results in Figure 9

|     | Gaussian noise | Salt and pepper noise | Poisson noise | Poisson noise |
| --- | --- | --- | --- | --- |
| DSC | 0.9582 | 0.9636 | 0.9493 | 0.9564 |
| JS  | 0.9541 | 0.9647 | 0.9472 | 0.9543 |

Plot the data in Table 2 into a line chart, as shown in Figure 10. From Figure 10, it can be observed that the segmentation performance of the model in this chapter is different for Gaussian noise, salt and pepper noise, Poisson noise, and speckle noise. The best performance is achieved when segmenting salt and pepper noise images, while the performance is slightly worse when segmenting Poisson noise images. This indicates that adding different levels of noise pollution to the image can also lead to significant differences in the boundaries of the image, This will result in different effects when the model grasps the object boundary. However, from the data in Table 2, it can be seen that for different types of noise, the calculated DSC and JS values can remain stable around 1, which proves that the model in this chapter is feasible for segmenting different types of noise and can accurately segment images.

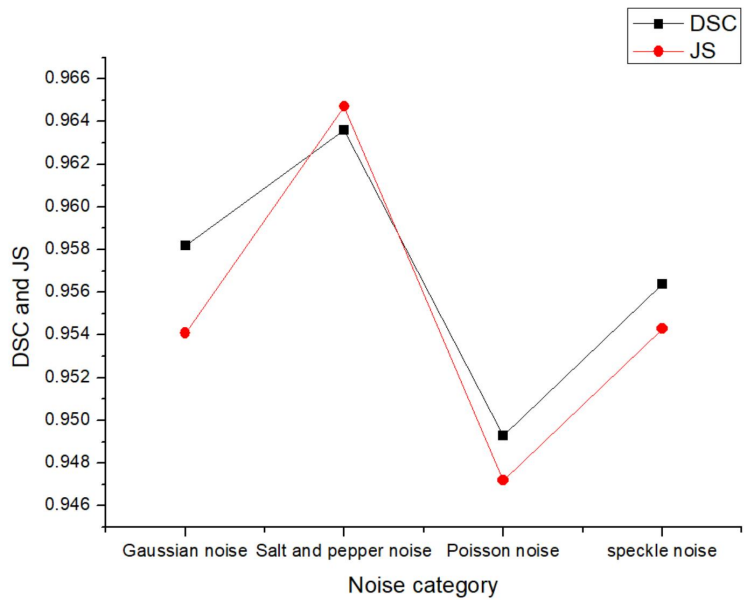

Figure 10 The broken line graph of DSC and JS values in Table 3.2

By segmenting four different types of noisy images and analyzing the two-step experiments in sequence, it was found that the HZSPF model has accurate segmentation results for different types and degrees of noisy images, which further demonstrates its good robustness to noise.

**4.3 Comparisons with the CV Model, the RSF Model, the WHRSPF Model and the SBGFRLS model**

(1)**Segmenting images with single-object intensity inhomogeneity.** As shown in Figure 11, the first image is the input image, the second image is the segmentation result of the CV model [9], the third image is the segmentation result of the RSF model [13], the fourth image is the segmentation result of the WHRSPF model [40], the fifth image is the segmentation result of the SBGFRLS model [38], and the last image is the segmentation result of the HZSPF model in this paper. It is evident from the figure that the CV model cannot segment the lower right corner of the palm, while the RSF model performs worse and the WHRSPF model performs better. However, the segmentation between the gap between the third and fourth fingers is still not complete enough. Similarly, the SBGFRLS model also has the same problem. Although the segmentation effect of the HZSPF model is not 100% accurate, it is much better than the other four models. The HZSPF model does not miss out on the details in the image and can relatively complete the segmentation of the image contour. This reflects the comparison with other models. The segmentation of intensity inhomogeneious images by the HZSPF model having its advantages.

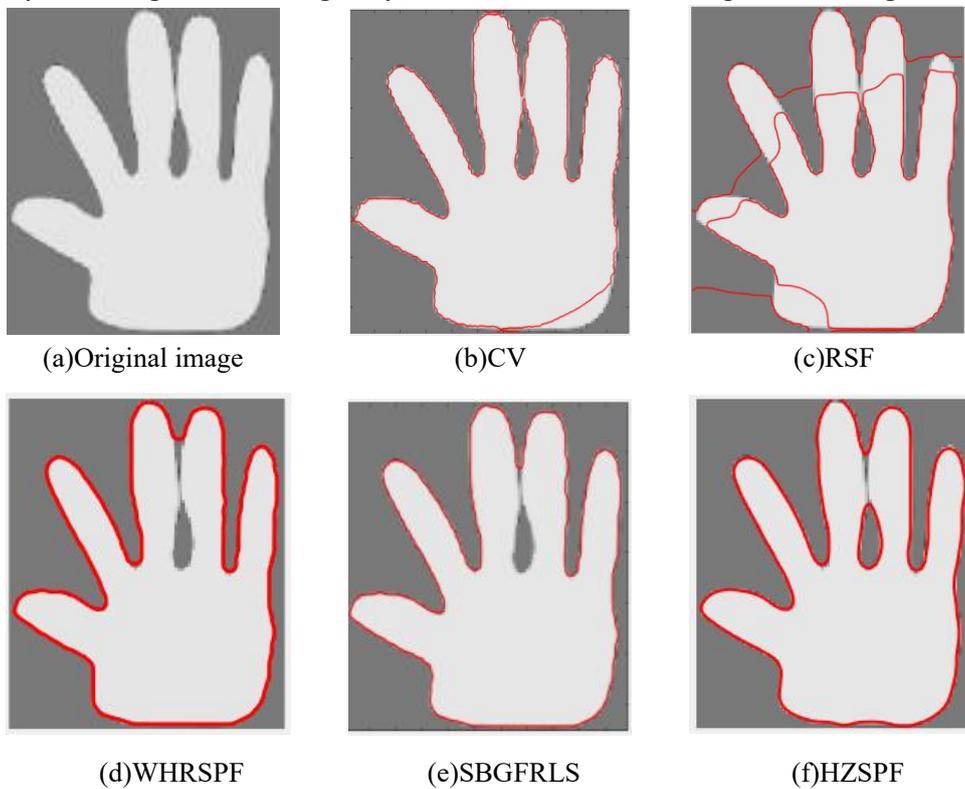

(a)Original image      (b)CV      (c)RSF

(d)WHRSPF      (e)SBGFRLS      (f)HZSPF

Figure 11 The segmentation results of the different model on the images with single-object intensity inhomogeneity

**(2)Segmenting complex medical images with intensity inhomogeneity.** As shown in Figure 12, the first image is the input image, the second image is the segmentation result of the CV model, the third image is the segmentation result of the RSF model, the fourth image is the segmentation result of the WHRSPF model, the fifth image is the segmentation result of the SBGFRLS model, and the last image is the segmentation result of the HZSPF model in this paper. It can be clearly seen from the results that the CV model, RSF model, WHRSPF model, and SBGFRLS model will misclassify some details in this complex intensity inhomogeneious image, resulting in incomplete segmentation. However, the HZSPF model clearly segments both the internal details and boundaries of the image, indicating that even when facing complex intensity inhomogeneious images, The HZSPF model is also relatively stable compared to other models.

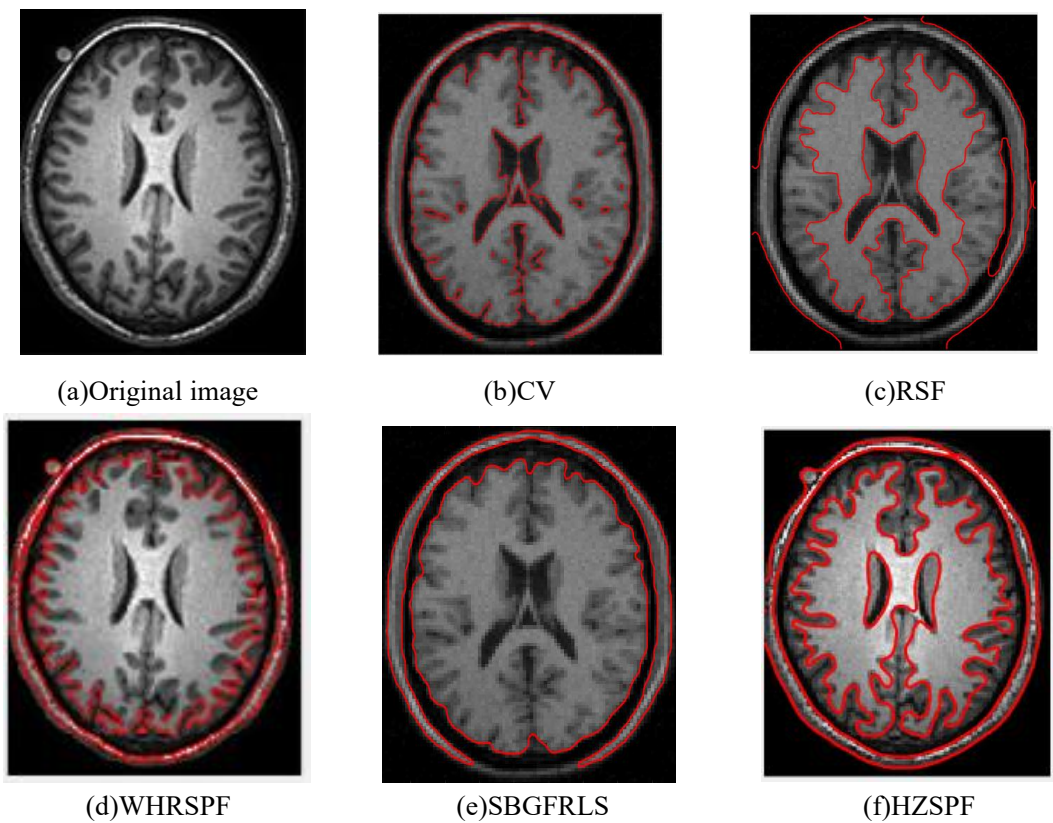

(a)Original image　　　　　　(b)CV　　　　　　(c)RSF

(d)WHRSPF　　　　　　(e)SBGFRLS　　　　　　(f)HZSPF

Figure 12 The segmentation results of the different model on the medical images

**(3)Segmenting images with multi-object intensity inhomogeneity.** As shown in Figure 13, the first image is the input image, the second image is the segmentation result of the CV model, the third image is the segmentation result of the RSF model, the fourth image is the segmentation result of the WHRSPF model, the fifth image is the segmentation result of the SBGFRLS model, and the last image is the segmentation result of the HZSPF model. It is evident from the results that the CV model, RSF model, and HZSPF model have similar segmentation effects and can accurately segment three objects. However, the WHRSPF model has poor segmentation performance at the object boundary, and the SBGFRLS model can only segment two objects. This also fully demonstrates that the HZSPF model has a significant advantage in segmenting multi-object intensity inhomogeneious images compared to similar WHRSPF and SBGFRLS models.

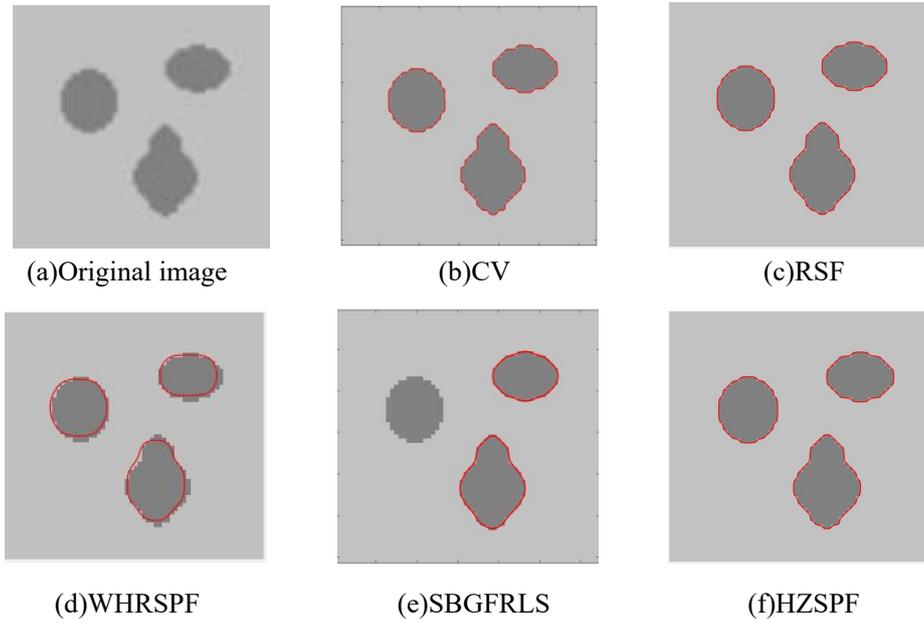

(a)Original image　　　　(b)CV　　　　(c)RSF

(d)WHRSPF　　　　(e)SBGFRLS　　　　(f)HZSPF

Figure 13 The segmentation results of the different model on the images with multi-object intensity inhomogeneity

Table 3 shows the DSC and JS values calculated for each model segmentation in Figures 11, 12, and 13 using equations (30), (31), and the groundtruth image. The results of DSC and JS show that compared to the other four models, the DSC and JS values of the model proposed in this chapter are closer to 1, indicating that the segmentation performance of the HZSPF model is better.

Table 3 DSC and JS values of segmentation results in Figure 11、Figure 12 and Figure 13

|  | CV | RSF | WHRSPF | SBGFRLS | MZSPF |
| --- | --- | --- | --- | --- | --- |
| Figure 11 | 0.8572 | 0.6728 | 0.9013 | 0.9346 | 0.9568 |
|  | 0.8461 | 0.6745 | 0.8941 | 0.9418 | 0.9573 |
| Figure 12 | 0.8673 | 0.6862 | 0.7124 | 0.7541 | 0.9675 |
|  | 0.8614 | 0.6735 | 0.7087 | 0.7426 | 0.9713 |
| Figure 13 | 0.9885 | 0.9861 | 0.9122 | 0.8426 | 0.9881 |
|  | 0.9913 | 0.9872 | 0.9087 | 0.8335 | 0.9927 |

# 5. Conclusion

In this paper, an active contour (HZSPF) model driven by hybrid signed pressure function is proposed for the segmentation of images with intensity inhomogeneity and noise. Firstly, by combining the average intensity of two internal and external regions with the median intensity of the internal region, a signed pressure function based on global information is introduced, which takes into account the global information of the image. Then design a signed pressure function that includes local information of the image. The total sign pressure function is obtained by linearly convex combining the sign pressure function containing global information and the sign pressure function containing local information. By substituting it into the level set evolution equation of the SBGFRLS model, we can obtain the Active Contour (HZSPF) model for image segmentation driven by a hybrid signed pressure function.

Finally, simulation experiments were conducted on the segmentation of intensity inhomogeneious images and noisy images using the HZSPF model, as well as comparing it with four similar models. In addition, quantitative analysis is also conducted to analyze the model in this paper based on the results of DSC coefficient and JS coefficient. Verified that the model in this paper has excellent robustness against intensity inhomogeneity and noise.